\documentclass[letterpaper]{article} 
\usepackage[]{aaai2026}  
\usepackage{times}  
\usepackage{helvet}  
\usepackage{courier}  
\usepackage[hyphens]{url}  
\usepackage{graphicx} 
\usepackage{natbib}  
\usepackage{caption} 
\usepackage{algorithm}
\usepackage{algorithmic}

\usepackage{newfloat}
\usepackage{listings}
\DeclareCaptionStyle{ruled}{labelfont=normalfont,labelsep=colon,strut=off} 
\floatstyle{ruled}
\newfloat{listing}{tb}{lst}{}
\floatname{listing}{Listing}

\usepackage[utf8]{inputenc} 
\usepackage[T1]{fontenc}    
\usepackage{url}            
\usepackage{booktabs}       
\usepackage{amsfonts}       
\usepackage{nicefrac}       
\usepackage{microtype}      
\usepackage{xcolor}         
\usepackage{graphicx}
\usepackage{amsmath}
\usepackage{amssymb}
\usepackage{amsthm}
\usepackage{mathrsfs}
\usepackage{float}
\usepackage{cuted}
\usepackage{natbib}
\usepackage{tcolorbox}
\usepackage{caption}
\tcbuselibrary{skins, breakable, listings, theorems}
\usepackage{silence}
\usepackage{enumitem}

\tcbset{
  promptbox/.style={
    enhanced,
    breakable,
    colback=gray!5, 
    colframe=black, 
    boxrule=0.8pt,
    arc=3pt,
    fonttitle=\bfseries\color{white}, 
    left=2pt,
    right=2pt,
    top=4pt,
    bottom=4pt,
    coltitle=white, 
    sharp corners=south,
    title={User Prompt},
    attach boxed title to top left={xshift=2pt, yshift=-2pt}, 
    title style={
      enhanced,
      frame code={\draw[black, line width=0.6mm] (frame.south west) rectangle (frame.north east);}, 
      interior code={\fill[black];} 
    }
  }
}

\tcbset{ 
    quotebubble/.style={
        enhanced,
        breakable,
        colback=blue!20,  
        colframe=blue,    
        boxrule=0.8pt,    
        arc=15pt,         
        width=0.95\textwidth,  
        top=4pt, bottom=4pt, left=4pt, right=4pt,  
        coltitle=white,
        fonttitle=\bfseries\color{white},  
        title={LLM Response},  
        attach boxed title to top left={xshift=2pt, yshift=-2pt},  
    }
}

\tcbset{ 
    imgllmbox/.style={
        enhanced, 
        breakable,
        colback=blue!10,  
        colframe=blue,    
        boxrule=0.8pt,    
        arc=3pt,          
        fonttitle=\bfseries\color{white},  
        left=2pt, right=2pt, top=4pt, bottom=4pt,
        coltitle=white,   
        title={Image Input \& LLM Response},  
        attach boxed title to top left={xshift=2pt, yshift=-2pt},  
        title style={ 
            enhanced, 
            frame code={\draw[blue, line width=0.6mm] (frame.south west) rectangle (frame.north east);},  
            interior code={\fill[blue];}  
        },
        colbacktitle=blue!30,  
    },
}

\tcbset{
    simplesummary/.style={
        enhanced,
        breakable,
        colback=orange!15,
        colframe=orange!80,
        boxrule=1pt,
        arc=3pt,
        left=4pt, right=4pt, top=6pt, bottom=6pt
    }
}

\title{The Forecast Critic: Leveraging Large Language Models for \\Poor Forecast Identification}

\author{
    Luke Bhan\textsuperscript{\rm 1}\thanks{Work completed at Amazon}, 
    Hanyu Zhang \textsuperscript{\rm 2}, 
    Andrew Gordon Wilson \textsuperscript{\rm 2},
    Michael W. Mahoney \textsuperscript{\rm 2}, 
    Chuck Arvin \textsuperscript{\rm 2}\thanks{Correspondence to chuarvin@amazon.com}
}
\affiliations{
    \textsuperscript{\rm 1}University of California, San Diego
    \textsuperscript{\rm 2}Amazon 
}

\begin{document}

\maketitle

\begin{abstract}
Monitoring forecasting systems is critical for customer satisfaction, profitability, and operational efficiency in large-scale retail businesses. 
We propose \texttt{The Forecast Critic}, a system that leverages Large Language Models (LLMs) for automated forecast monitoring, taking advantage of their broad world knowledge and strong ``reasoning'' capabilities. 
As a prerequisite for this, we systematically evaluate the ability of LLMs to assess time series forecast quality, focusing on three key questions. 
(1) Can LLMs be deployed to perform forecast monitoring and identify obviously unreasonable forecasts? 
(2) Can LLMs effectively incorporate unstructured exogenous features to assess what a reasonable forecast looks like?
(3) How does performance vary across model sizes and reasoning capabilities, measured across state-of-the-art LLMs?
We present three experiments, including on both synthetic and real-world forecasting data. Our results show that LLMs can reliably detect and critique poor forecasts, such as those plagued by temporal misalignment, trend inconsistencies, and spike errors. 
The best-performing model we evaluated achieves an F1 score of $0.88$, somewhat below human-level performance (F1 score: $0.97$). 
We also demonstrate that multi-modal LLMs can effectively incorporate unstructured contextual signals to refine their assessment of the forecast. 
Models correctly identify missing or spurious promotional spikes when provided with historical context about past promotions (F1 score: $0.84$). 
Lastly, we demonstrate that these techniques succeed in identifying inaccurate forecasts on the real-world M5 time series dataset, with unreasonable forecasts having an sCRPS at least 10\% higher than that of reasonable forecasts.
These findings suggest that LLMs, even without domain-specific fine-tuning, may provide a viable and scalable option for automated forecast monitoring and evaluation. 
\end{abstract}

\section{Introduction} 
\label{sec:introduction}

With the rise of large-scale distributed computing, automated purchasing systems that manage inventory decisions at the scale of millions of products are becoming increasingly common. 
Such purchasing systems require accurate predictions of product demand at all times to make optimal decisions. 
However, designing accurate forecasters at scale is challenging for a variety of reasons. 
For example, exogenous shocks like weather events or promotional events can skew observed values, and the inductive biases of time series models may lead to a failure to preserve important time series characteristics, like trends, periodicity, or volatility \cite{Yu2025UnderstandingTransformersTS,Yu2025UnderstandingImplicitBiases}.
To address these challenges, time series foundation models \cite{ansari2024chronos, 10.5555/3692070.3692712, 10.5555/3666122.3666983, 10.5555/3692070.3694248} have gained attention, delivering strong cross-timescale performance across domains such as weather prediction and demand forecasting, often without retraining. 
Even as models produce strong results in aggregate backtests when averaged across a full dataset, these models can sometimes produce poor individual forecasts, with obvious errors like misaligned trends, unrealistic transitions, or missing periodicity. Further, even models which perform well in backtests may produce unrealistic forecasts in the real world, where input data corruption or model deployment errors may trigger unanticipated behavior.

Consequently, to deploy such models in mission-critical real-world applications, we require systems that can reliably identify such inaccurate forecasts without knowing the true future observations. 
Human experts can often make such judgments by \emph{visually} inspecting historical patterns and assessing whether a forecast is \emph{plausible}, given the observed context. 
Inspired by this, we design a Large Language Model (LLM) based system that emulates such a qualitative visual human plausibility evaluation, enabling automated assessment of time series forecasts directly from visual plots.
We call this system \texttt{The Forecast Critic}.

\texttt{The Forecast Critic} is designed to interpret a plot of a historical time series, unstructured guidance about what a reasonable forecast looks like, and the corresponding forecast, to determine whether the forecast is plausible or reasonable. 
Beyond serving as a human-like evaluator, such an approach offers multiple advantages for validating forecasts as they are generated, rather than in hindsight. 
First, to the extent that LLM behavior parallels human reasoning, the model may automatically detect key features of a time series, including trends, periodic patterns, and volatility levels. 
This is advantageous over statistical approaches that require domain-based hand-tuning from an engineer and that may incorrectly weigh certain characteristics of the time series over others.
Second, LLMs can operate as automated processes, allowing them to process many more time series than would be possible if the process relied on human experts to conduct this forecast auditing.
Lastly, multi-modal LLMs can integrate additional contextual information, including instructions or exogenous features, without retraining. 
This additional context allows the models to incorporate new signals in an online manner, layering on additional information to foundation models trained on increasingly large datasets \cite{wolff2024}.

To illustrate the kinds of errors that such an automated LLM agent can identify, we showcase in Figure \ref{fig:m5-example} a real-world forecast from the Chronos model on the M5 dataset. 
In this example, the LLM correctly identifies that the forecast fails to reflect the high volatility present in the historical time series, and thus it identifies that this forecast is unreasonable. 
This example highlights the potential of this technology for practical deployment at scale. 

Building on this illustrative example, we first evaluate \texttt{The Forecast Critic} system on controlled experiments with univariate synthetic data, where we artificially introduce standard but easily-identified perturbations by modifying the forecast trend, translating the forecast up or down, compressing the periodicity of the forecast, and introducing random outlier spikes to the forecasts.
Such perturbations are relatively easy for human experts to identify (F1 score: $0.97$). 
We find that LLMs are generally reasonably capable of automatically identifying these unreasonable forecasts, with the best-performing LLM achieving an F1 score of $0.88$. 
We then assess how well these LLM-based forecast critiques can incorporate additional contextual information by providing the LLM with spurious and meaningful information about upcoming promotional events.
We demonstrate that LLMs are correctly able to reason about such features, providing a promising avenue for including new features in an online manner through \texttt{The Forecast Critic}. 
Lastly, we conclude by evaluating performance on real-world time series from the M5 dataset \cite{MAKRIDAKIS20221346}, using forecasts generated by Chronos at the product granularity. 
Using this real-world validation, we show that forecasts flagged by \texttt{The Forecast Critic} have significantly worse error rates than unflagged forecasts.

\begin{strip}
    \centering
    \begin{minipage}{\textwidth}
    \begin{tcolorbox}[imgllmbox]
    \begin{minipage}{0.48\textwidth}
    \centering
    \includegraphics[width=0.7\textwidth]{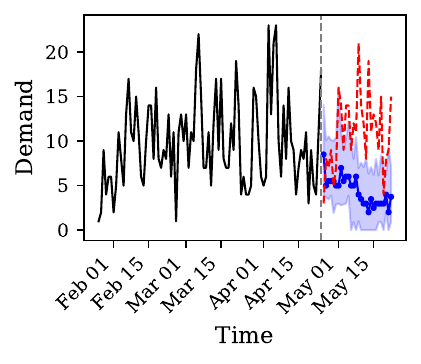} \ \\
    \includegraphics[width=\linewidth]{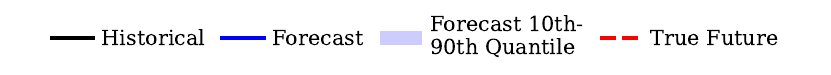}
    \end{minipage}
    \hspace{5pt} 
    \begin{minipage}{0.48\textwidth}
    \begin{tcolorbox}[quotebubble]
    \small{Looking at this image, I can see historical data showing high volatility with frequent spikes reaching values up to around 20. The forecast appears much less volatile, projecting values mostly between 3-7 with minimal fluctuation compared to the historical pattern. 
    
    This forecast fails to capture the significant volatility that is a consistent characteristic of the historical data. The drastic reduction in volatility represents a fundamental mismatch with the historical behavior of this time series, as there's no evidence that volatility was decreasing prior to the forecast point.}
    \end{tcolorbox}
    \end{minipage}
    \centering 
    \vspace{0.5em}

\end{tcolorbox}
\captionof{figure}{Example of a flagged forecast and the corresponding model justification for the M5 dataset using Claude 3.7 Sonnet. The model is fed the image on the left without the future values (shown in red), along with a prompt (given in the M5 section, below). The LLM flags this forecast as unreasonable, as it ``fails to capture the significant volatility'' in the historical time series.}
\label{fig:m5-example}
\end{minipage}
\end{strip}

\subsection{Related work}

\noindent\textbf{LLMs as tools in retail applications.}
LLMs have been deployed across a variety of retail applications, including marketing \cite{10.1145/3589335.3651520}, product reviews \cite{ROUMELIOTIS2024100056}, product search \cite{10.1145/3642979.3643007}, and personalized data visualization \cite{10.1145/3701716.3715452}. 
In particular, they show promise in zero-shot or few-shot contexts, allowing them to reason about novel tasks and situations, without requiring retraining \cite{10.1145/3664647.3681190}, achieving performance levels similar to those of human experts \cite{chen2023knowledgegraphcompletionmodels}. 
LLMs have also been commonly investigated as chat agents, improving sales \cite{liu2024llmconversationalagentmemory} and customer service retention \cite{pandya2023automatingcustomerserviceusing}, and they have even been used as customer relationship management (CRM) systems \cite{huang-etal-2025-crmarena}. 
Despite initial successes, significant challenges remain in the systematic deployment of LLMs at scale in retail settings \cite{ren2024surveyfairnesslargelanguage}. 
In this work, we highlight these challenges through a critical examination of time series forecasting.

\noindent\textbf{LLMs for augmenting time series forecasting.} 
Beyond these general retail applications, LLMs have garnered significant attention as powerful tools for enhancing time series forecasting frameworks. 
These models have shown promising performance across a variety of forecasting tasks \cite{10.5555/3666122.3666983}, though debate remains regarding their effectiveness in zero-shot deployment scenarios \cite{tan2024are}. 
Importantly, LLMs are not limited to simply generating forecasts.
They have also been used to refine baseline forecasts by incorporating key external events like promotions or holiday events that may influence the time series \cite{ashok2025naivepromptingstrategiesimproved, zhang2024llmforecasterimprovingseasonalevent}. 
LLMs have also proven valuable in detecting anomalies within time series data, often by interpreting visual representations of the time series \cite{zhou2025can, xu2025multimodalllmsperformtime}. 
These models are adept at analyzing and interpreting a wide range of time series characteristics, enabling deeper insights into time series behaviors and underlying dynamics \cite{fons-etal-2024-evaluating, cai2024timeseriesexamtimeseriesunderstanding}. 
Finally, LLMs have been effectively applied to the problem of forecasting model selection. 
By processing extracted time series features, they can recommend suitable forecasting models and fine-tuned hyperparameters \cite{wei2025efficientmodelselectiontime}. 

Despite these advancements, limited work has explored the combined challenges of forecast monitoring, which requires an LLM to understand the historical time series and evaluate whether a forecast is a reasonable extrapolation of that history. In this work, we address this gap by focusing on both aspects, with particular attention to developing automated tools to distinguish between accurate and inaccurate (or plausible versus implausible) forecasts.

\section{Can LLMs detect synthetic time series perturbations?} \label{sec:synthetic}

In this section, we address the following questions. 
(1) Can LLMs detect common forecasting issues from visual plots alone? 
(2) Are certain issues more difficult for LLMs to detect than others? 
(3) Can LLMs incorporate exogenous information to flag missing or spurious promotional spikes?
We demonstrate that state-of-the-art LLM models can correctly identify when a forecast is translated, contains false spikes, exhibits an unreasonable trend magnitude, or is stretched/compressed, relative to the true time series. 
LLM performance is comparable, though somewhat worse, than that of an expert human. 
Further, we show that performance varies depending on the perturbation introduced. 
Additionally, we highlight that multi-modal LLMs are successfully able to connect text containing exogenous promotional information with the forecast to improve contextual decision making.

\subsection{Dataset construction} 
\label{sec:subsec:dataset}

As demonstrated in \cite{dooley2023forecastpfn} and \cite{taga2025timepfn}, synthetic time series are effective for training real-world forecasting models. 
To begin our systematic study of \texttt{The Forecast Critic}, we first apply simple perturbations to synthetically generated time series in a point-wise prediction setting. 
This approach offers two key advantages over the real-world experiments presented later in this paper. 
First, it allows us to evaluate the performance of LLMs across a variety of time series errors, helping identify potential situations where LLMs would miss inaccurate forecasts in a real-world deployment. 
Second, synthetic data provides a clear benchmark for determining when a forecast should be considered inaccurate. 
In real-world applications, the definition of a ``good'' forecast may be ambiguous, depending on both the time series and the preferences of forecast users. 
In the synthetic setting, by contrast, we control the data-generating process, so identifying good versus bad forecasts is more straightforward.

\noindent
\textbf{Synthetic time series generation.}
We construct a synthetic dataset consisting of various basis functions (see the Appendix for dataset details). 
Let  $\mathcal{T} = \{t_0 + k \cdot \Delta t \mid k = 0, 1, \dots, N\}$ be a set of discrete evaluation points, where \(\Delta t\) is the sampling interval, and the series is divided into historical data  \(t \in \mathcal{T}_{\text{historical}}\) and forecast data \(t \in \mathcal{T}_{\text{forecast}}\) such that $\mathcal{T} = \mathcal{T}_{\text{forecast}} \cup \mathcal{T}_{\text{historical}}$. 
Then, we define the time series $y(t)$~as 
\begin{align}
    y(t) = \sum_{i=1}^n w_i \cdot b_i(s_i(t + \delta_i))\,,  \label{eq:ts-main} \quad t \in \mathcal{T}\,, 
\end{align}
where $n \sim \mathcal{U}\{1, 2, 3, 4\}$ is the number of basis functions to include, and for all $i \in [1, n]$, we have $w_i \sim \mathcal{U}(0.5, 2.0)$ as an output scaling weight, $s_i \sim \mathcal{U}(0.5, 2.0)$ as an input scaling weight, $\delta_i \sim \mathcal{U}(0, 4)$ as an input shift, and $b_i \sim \mathcal{U}\{1, 2, ..., 14\}$ as the basis functions defining the time series (given in the Appendix). 
For all the experiments, we consider a $\Delta t=0.1$ and a final time horizon of $T=10$ (yielding $100$ discrete points), and we use the time $t=8$ as the split between the historical window and forecasting period.

\vspace{2mm}
\begin{strip}
\begin{tcolorbox}[promptbox]
\begin{minipage}{0.48\textwidth}
\small{You are shown an image of historical data (in black) and a forecast (in blue). 
Based on the historical trend, assess whether the forecast is reasonable. A reasonable forecast should generally follow the same direction and capture any seasonal trends if there are any. 

Please provide a brief explanation (1–2 sentences) justifying your decision. 
Then present your final answer using one of the following options, wrapped in <answer> tags:

<answer> 1 </answer> — The forecast is reasonable.  

<answer> 2 </answer> — The forecast is unreasonable.}
\end{minipage}%
\hfill
\begin{minipage}{0.48\textwidth}
\centering
\includegraphics[width=0.7\linewidth]{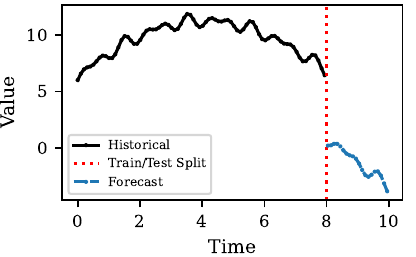}
\end{minipage}
\end{tcolorbox}
\noindent \begin{minipage}{\textwidth}
\captionof{figure}{Example prompt asking an LLM whether a forecast (featuring an artificially-induced downward translational shift) is reasonable or unreasonable. In this case, the LLM should respond with unreasonable, along with a justification noting that the forecast is unreasonably low.}
\label{fig:example_prompt}
\end{minipage}
\end{strip}

\noindent\textbf{Forecast perturbations.} 
We begin by synthetically injecting perturbations into the forecast data to create clearly inaccurate, or unreasonable,  
forecasts. 
As highlighted in previous works \cite{zhou2025can}, casting time series data as image inputs for multi-modal LLMs is a viable—and sometimes superior—approach for time series tasks. 
Consequently, we present our inputs to the LLMs in the form of plots, an example of which is shown in Figure \ref{fig:example_prompt}. 
We consider five different state-of-the-art LLMs that accept multi-modal text-image inputs and produce text outputs.
These models have varying reasoning capabilities and model sizes: Claude 3.7 Sonnet \cite{anthropic2025claude3.7}, Claude 4.0 Sonnet \cite{anthropic-opus4-system-card}, Claude 4.0 Sonnet with reasoning, Claude 4.0 Opus \cite{anthropic-opus4-system-card} and Meta Llama 4 Maverick \cite{meta_ai_Llama4}.

To evaluate the LLMs, we consider four distinct types of perturbations (see Figure \ref{fig:perturbations} and the Appendix 
for examples of each), and we construct case studies where half of the examples contain a  perturbation and the other half consist of unaltered, correct forecasts. 
Each model is prompted, as illustrated in Figure \ref{fig:example_prompt}, to assess whether a forecast is reasonable. 
We evaluate performance on 500 unique time series forecasts, equally split between perturbed and unperturbed cases (250 perturbed, 250 without modification). 
To ensure the perturbed examples are visibly worse than the true time series, we generate 334 problems per perturbation type, and we retain the worst 75\% based on symmetric mean absolute percentage error (SMAPE) to obtain our 250 examples of each perturbation. 
Each LLM is evaluated using its default settings via Amazon Bedrock, and we report both class-specific F1 scores and the overall weighted F1 score (see the Appendix 
for definitions) in Figure \ref{fig:synthetic-results-clean}. 
For comparison, we establish a human baseline by having an expert annotator independently label the same 500 plots using the same interface. 

\begin{figure}
    \centering
    \includegraphics[width=\linewidth]{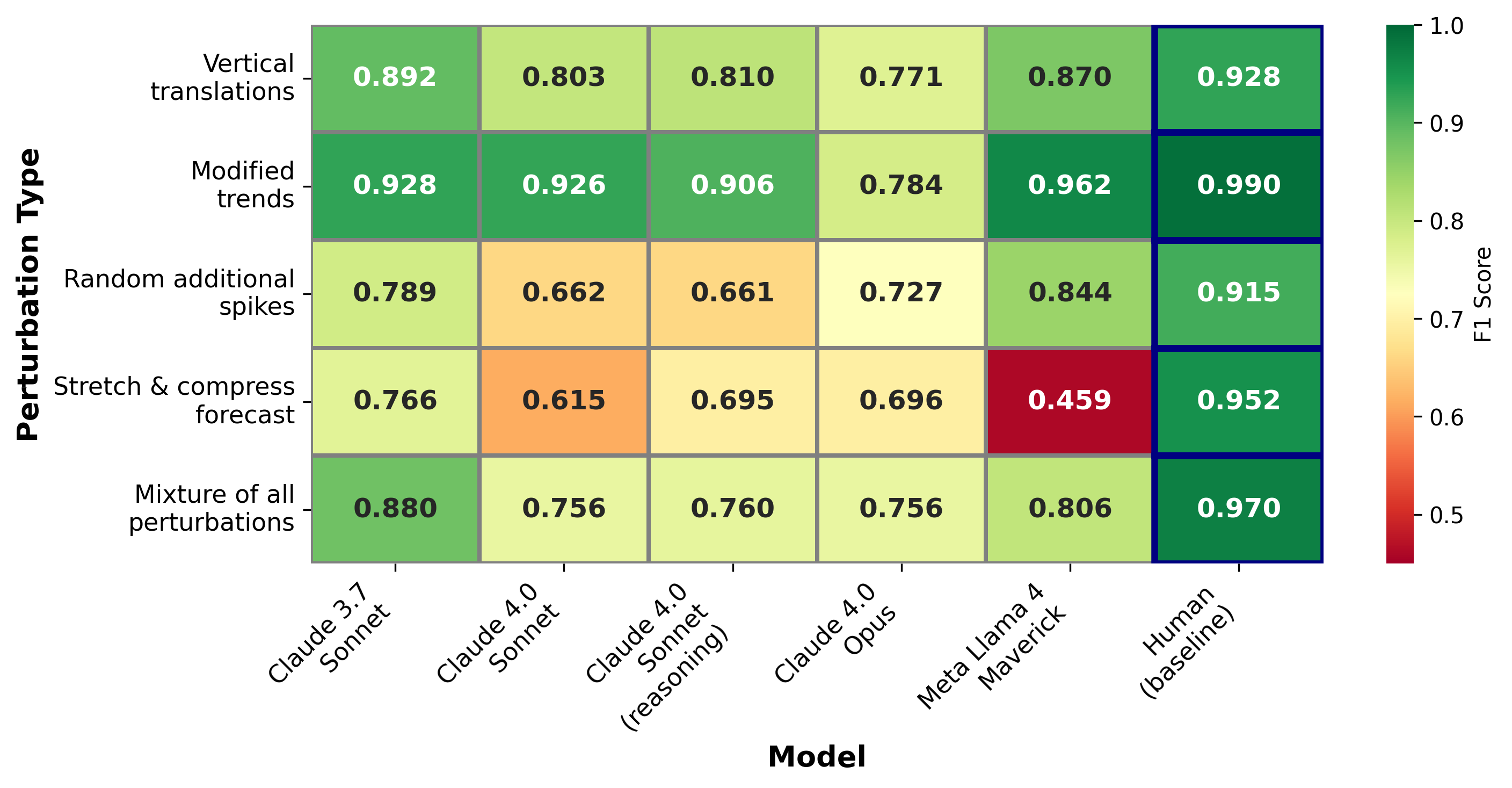}
    \caption{Evaluation of the ability of LLMs to identify reasonable versus unreasonable forecasts based on various synthetic perturbations.}
    \label{fig:synthetic-results-clean}
\end{figure}

\subsection{Empirical Results}

We find that model performance varies substantially depending on the type of perturbation applied.
See Figure \ref{fig:synthetic-results-clean} for a summary, and see Appendix Figures \ref{fig:perturb-example-1}, \ref{fig:perturb-example-2}, \ref{fig:perturb-example-3}, \ref{fig:perturb-example-4} for example LLM anecdotes. 
We find that, depending on perturbation, the Claude 3.7 Sonnet or Meta Llama 4 Maverick models perform best. 
This is notable, as both models are older than more recent counterparts like Claude 4.0 Sonnet and Opus. 
We find that LLMs are quite capable of identifying situations where the forecast has an unrealistic trend (F1 score $> 0.9$), and they are generally capable of identifying forecasts that have been incorrectly shifted or have spurious forecast spikes. 
By contrast, the LLMs struggle to identify cases where the forecasts have been unrealistically stretched or compressed (as in Figure \ref{fig:perturb-example-3}), even when human experts have little difficulty with those examples.
Additionally, we find that the human evaluator outperforms all LLM models in all categories, highlighting the room for improvement. 
In the final row of Figure \ref{fig:synthetic-results-clean}, we present results from a more difficult ``mixture'' evaluation, in which any of the four perturbations may be randomly applied. 
This setup helps to mitigate annotator bias by preventing the human experts from seeing several examples of a single perturbation in sequence, increasing task complexity. 
Despite this, both the human and Claude 3.7 Sonnet continues to perform well, achieving F1 scores of $0.97$ and $0.88$, respectively. 

These findings suggest that neither model size nor newer model generations reliably improve performance on \texttt{The Forecast Critic} task. These results are somewhat surprising.
While newer models like Claude 4.0 Opus are generally more capable, we see meaningful degradation on this particular task.
Overall, these results show that top performing LLMs are capable of reliably identifying inaccurate forecasts, at least for cases where the inaccuracy is fairly obvious by construction. 
While the top performing LLMs still underperform human experts, F1 scores on this task often exceed $0.8$, indicating a reliable ability to identify inaccurate predictions.

\begin{tcolorbox}[simplesummary]
    \textbf{Summary: } LLMs succeed at identifying clearly inaccurate forecasts, with F1 score approaching 0.9. LLMs also reliably identify modified trends and vertical translations, but they struggle with stretching and compressed periodicity.
\end{tcolorbox}

\subsection{Exogenous Feature Injection} 
\label{sec:subsec:exogenous-feature-injection}

A key advantage of using LLMs as forecast auditors lies in their ability to use unstructured context, including plain-language instructions or exogenous features in a zero-shot manner \cite{ghasemloo2025informedforecastingleveragingauxiliary}. 
In this experiment, we systematically examine how well LLMs can do this in the context of \texttt{The Forecast Critic}. 

We construct synthetic data reflecting a series of different promotional scenarios, and we evaluate the ability of LLMs to correctly reason about each of the following scenarios.
\begin{enumerate}[label=(\alph*)]
    \item Historical sales showed no lift during past promotions. There is an upcoming promotion, but there is no spike in the forecast. This is \emph{reasonable}, since the sales do not depend on the promotion.  
    \item Historical sales showed no lift during past promotions. There is an upcoming promotion, and there is a spike in the forecast corresponding to the upcoming promotion. This is \emph{unreasonable}, since the forecast expects a lift in sales that is not shown in the historical data.
    \item Historical sales showed a lift during past promotions. There is an upcoming promotion, but there is no spike in the forecast for the upcoming promotion. This is \emph{unreasonable}, since the forecast fails to anticipate a lift in sales from the upcoming promotion.
    \item Historical sales showed a lift during past promotions. There is an upcoming promotion, and there is a spike in the forecast for the upcoming promotion. This is \emph{reasonable}, since the forecast anticipates the upcoming promotional lift.
\end{enumerate}
Examples of each scenario are given in Appendix Figure \ref{fig:holiday-examples}. 
For each scenario, we evaluate the LLM on 500 examples modifying the synthetic dataset with the corresponding promotional spikes. 
We then ask the LLM whether the forecast is reasonable or unreasonable, using a prompt that includes the dates of the holidays. An example of the input and output is shown in Figure \ref{fig:text-exogenous-example-2}.

\begin{strip}
\centering
\begin{minipage}{\textwidth}
    \begin{tcolorbox}[imgllmbox]
        \begin{minipage}{0.48\textwidth}
        \centering
        \includegraphics[width=0.8\linewidth]{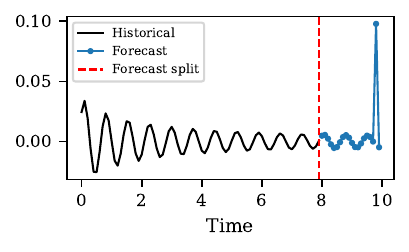} 
        \end{minipage}
    \begin{minipage}{0.48\textwidth}
    \begin{tcolorbox}[quotebubble]
    \small{Looking at the image, I can see that the historical data (black line) shows a clear cyclical pattern with decreasing amplitude over time. The forecast (blue line) maintains the cyclical pattern but doesn't continue the dampening trend, instead showing relatively stable oscillations. Additionally, the holiday spike at $t=9.82$ in the forecast appears significantly larger than the comparable holiday spike in the historical data at $t=1.88$, which doesn't align with the established pattern. These inconsistencies suggest the forecast doesn't properly capture the historical trend characteristics.}
    \end{tcolorbox}
    \end{minipage}
\end{tcolorbox}
\captionof{figure}{
Example of an LLM considering additional promotional context when evaluating the forecast. The LLM reasons that the forecast spike at $t=9.82$ is much larger than the promotional behavior in the history, so it flags the example as unreasonable.}
\label{fig:text-exogenous-example-2}
\end{minipage}
\end{strip}

\begin{figure}[t]
    \centering
    \includegraphics[width=0.95\linewidth]{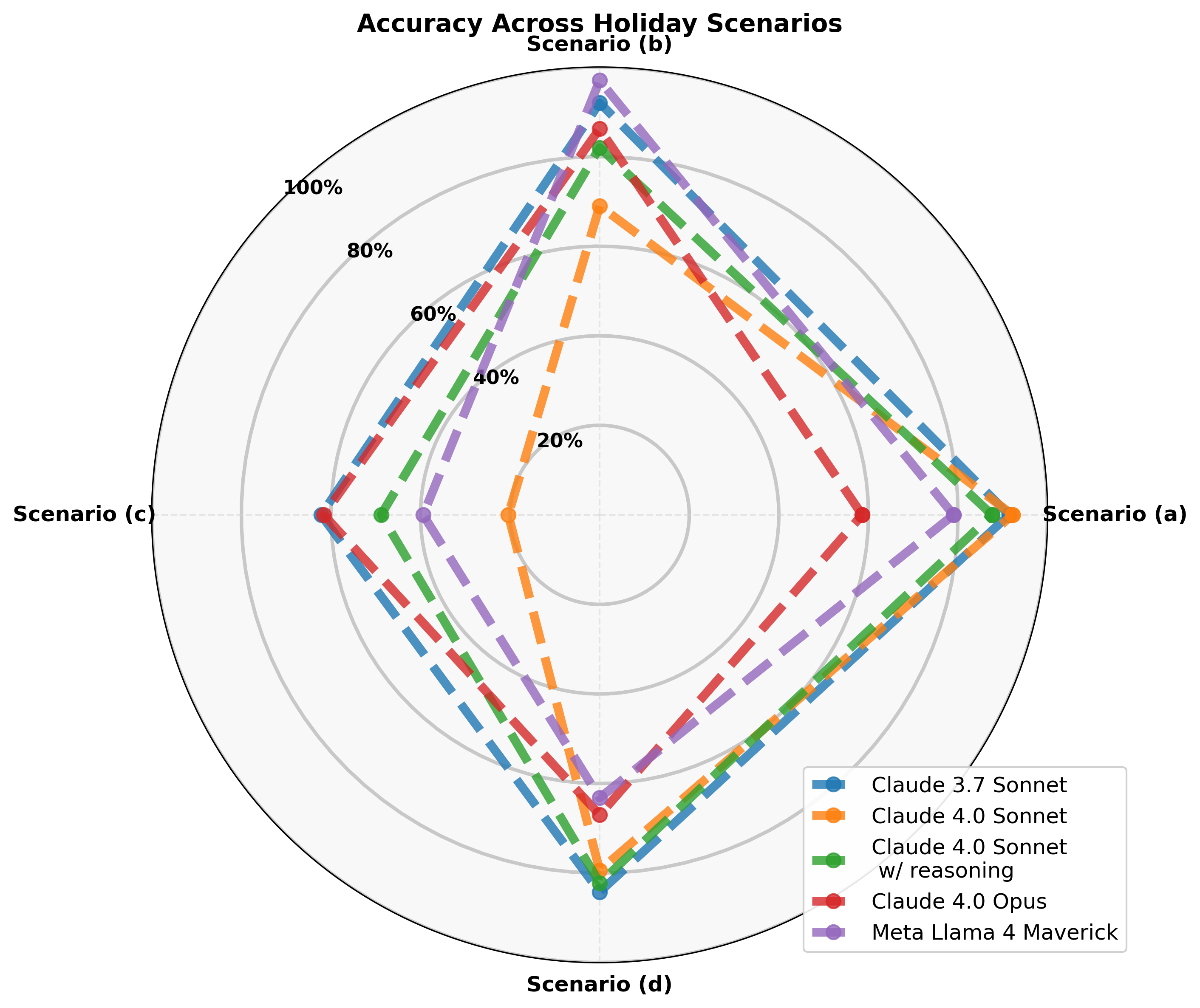}
    \caption{{LLM performance on exogenous feature injection across four promotional holiday scenarios. Scenario (a): product sales do not depend on promotions, with no spikes shown in either history or forecast (reasonable forecast). Scenario (b): product sales do not depend on promotions, but forecast incorrectly shows a spike (unreasonable forecast). Scenario (c): product sales depend on promotions with spike in history, but forecast is missing the spike (unreasonable forecast). Scenario (d): product sales depend on promotions, with spikes correctly shown in both history and forecast (reasonable forecast). Accuracy represents the percentage of forecasts correctly classified as reasonable or unreasonable.  Claude 3.7 Sonnet achieves the best overall performance with an F1-score of 0.836, followed by Claude 4.0 Sonnet with reasoning (F1: 0.767) and Meta Llama 4 Maverick (F1: 0.706).}
    }
    \label{fig:tab:holiday}
\end{figure}

We present a summary of the results in Figure \ref{fig:tab:holiday}, where we see that across all scenarios, the best performing model (Claude 3.7 Sonnet) achieves F1-scores of around $0.8$. Here again, newer versions of the Claude model show degradation in performance on this task.
Notably, we also see that scenario (c) is the hardest task across all models tested. 
In scenario (c), the forecasts are missing a promotional spike, even though such a spike is observed in the historical sales. This result is especially concerning, as detecting cases where a model fails to anticipate an upcoming promotion is a critical use case for such a forecast inspection system.
We provide additional illustrative examples of the generated analyses for all scenarios in Appendix Figures \ref{fig:exogenous-example-1}, \ref{fig:exogenous-example-2}, \ref{fig:exogenous-example-3}, and \ref{fig:exogenous-example-4}. 

\begin{tcolorbox}[simplesummary]
    \textbf{Summary: } LLMs can incorporate additional exogenous information to determine whether a forecast is reasonable, with F1 score up to 0.84. All models struggle most with scenario (c), where the forecast is missing a promotional spike, despite seeing elevated sales in the history.
\end{tcolorbox}

\section{Can LLMs identify poor real-world forecasts from foundation models?} 
\label{sec:m5}

\subsection{M5 Dataset construction} 
\label{sec:subsec:m5-dataset}

While synthetic data is useful for systematic studies of LLMs, they do not fully reflect real-world performance. To bridge this gap, we conduct an additional study to assess whether LLMs can correctly flag inaccurate forecasts from the Chronos model \cite{ansari2024chronos} using the M5 time series competition dataset \cite{MAKRIDAKIS20221346}. The M5 challenge involves hierarchical forecasting, where participants predict demand for various products across multiple levels, including states, departments, and categories. We consider the product aggregation level, and we randomly sample 1,000 time series and corresponding forecasts.
The time series is at the daily granularity, and we present the LLM with 120 days of historical time series and 28 days of forecasts. Because Chronos provides a full distribution forecast, we modify the figures to include the median, 10th and 90th percentiles of the distribution, as shown in Figure~\ref{fig:m5-example}. We present the LLM only a univariate view of the time series, without additional covariates and contextual~data. 

\subsection{Empirical results} 
\begin{figure}[h]
    \centering
    \includegraphics[width=\linewidth]{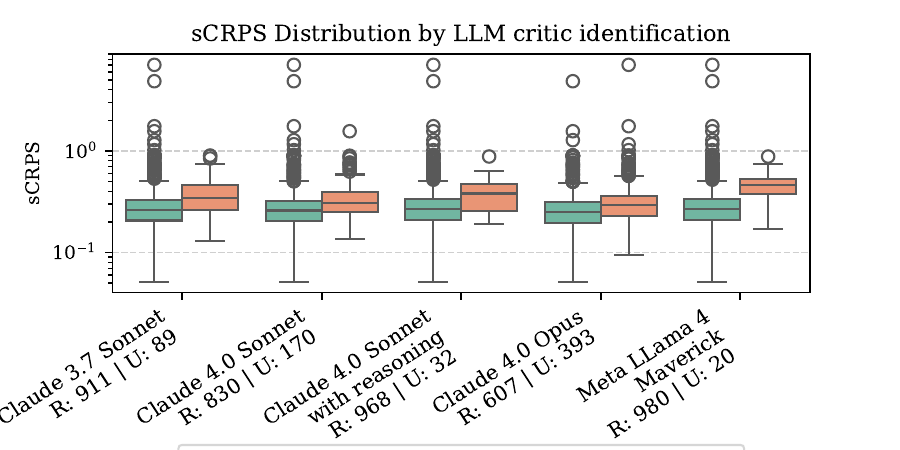}
    \caption{Comparison of sCRPS distributions on the forecasts flagged as reasonable and unreasonable for each critic model. 
    Here, $R:X|U: Y$ stands for the number of reasonable and unreasonable forecasts flagged by the model. Claude 3.7 Sonnet assessed 91.1\% of the forecasts as reasonable, while Claude 4.0 Opus assessed only 60.7\% of the forecasts as reasonable.}
    \label{fig:distributions-m5}
\end{figure}

Designing a rigorous experiment to evaluate real-world forecasts introduces a core challenge: defining what makes a forecast \emph{good} is inherently ambiguous. 
To help navigate this, we use the scaled continuously ranked probability score (sCRPS) (see the Appendix 
for definitions) as our evaluation metric, with higher scores indicating less accurate forecasts. Importantly, sCRPS requires access to ground-truth future values, and it cannot be validated \emph{a priori}, but it provides a viable method for systematically evaluating whether \texttt{The Forecast Critic} identified forecasts which were going to be inaccurate. Additionally, in initial experiments we observed that the models considered the vast majority of forecasts unreasonable. To address this, we provided additional instructions to the LLM instructing it to only flag \emph{obvious} forecasting errors, leaving the interpretation of obvious largely to the LLM (see Appendix Figure~\ref{fig:m5_example_prompt}).

Under this setup, several key patterns emerge. First, different models are much more likely to flag a forecast as unreasonable, ranging from just 2\% for Llama 4 Maverick to nearly 40\% for Claude 4.0 Opus. Despite this, forecasts flagged by the LLM as unreasonable consistently have higher sCRPS values than reasonable ones, as shown in Figure~\ref{fig:distributions-m5}. Across all models, the distribution of sCRPS for reasonable and unreasonable forecasts are statistically distinct according to a Mann-Whitney U test.
See Table~\ref{tab:m5-agg} for details.
Further, the median sCRPS for unreasonable forecasts is at least 10\% higher than for reasonable ones. Third, there is clear room to further improve the LLM performance, as Figure \ref{fig:distributions-m5} shows that all models consider some forecasts with very high sCRPS to be reasonable. Finally, we present anecdotal examples (Appendix Figures~\ref{fig:m5-ad-2} and \ref{fig:m5-ad-3}), including a failure mode where an LLM incorrectly flags a reasonable forecast due to an outlier spike in the historical data. Despite such individual errors, the aggregate results demonstrate that \texttt{The Forecast Critic} is effective at identifying forecasts that have significantly worse accuracy.

\begin{table*}[]
\centering
\resizebox{0.75\textwidth}{!}{%
\begin{tabular}{l|llllll}
Model &
  \begin{tabular}[c]{@{}l@{}}Claude 3.7\\ Sonnet\end{tabular} &
  \begin{tabular}[c]{@{}l@{}}Claude 4.0\\ Sonnet\end{tabular} &
  \begin{tabular}[c]{@{}l@{}}Claude 4.0\\ Sonnet\\ with reasoning\end{tabular} &
  \begin{tabular}[c]{@{}l@{}}Claude 4.0 \\ Opus\end{tabular} &
  \begin{tabular}[c]{@{}l@{}}Meta Llama\\ 4 Maverick\end{tabular} \\ \hline
\begin{tabular}[c]{@{}l@{}}Reasonable,  Median\end{tabular} &
  0.2624 &
  0.2591 &
  0.2661 &
  0.2505 &
  0.2661 \\
\begin{tabular}[c]{@{}l@{}}Reasonable,  Mean\end{tabular} &
  0.3022 &
  0.3008 &
  0.3063 &
  0.2505 &
  0.2661 \\
\begin{tabular}[c]{@{}l@{}}Reasonable, Standard  Deviation\end{tabular} &
  0.3105 &
  0.3206 &
  0.3044 &
  0.24 &
  0.3022 \\ \hline
\begin{tabular}[c]{@{}l@{}}Unreasonable,  Median\end{tabular} &
  0.3417 &
  0.3081 &
  0.3804 &
  0.2924 &
  0.4622 \\
\begin{tabular}[c]{@{}l@{}}Unreasonable,  Mean\end{tabular} &
  0.3752 &
  0.3472 &
  0.3813 &
  0.3403 &
  0.4658 \\
\begin{tabular}[c]{@{}l@{}}Unreasonable,  Standard  Deviation\end{tabular} &
  0.1615 &
  0.1704 &
  0.1479 &
  0.3739 &
  0.168 \\ \hline
\begin{tabular}[c]{@{}l@{}}Median, \% Diff\end{tabular} &
  30.2 &
  18.9 &
  43 &
  16.7 &
  73.7 \\
\begin{tabular}[c]{@{}l@{}}Mean, \% Diff\end{tabular} &
  24.18 &
  15.45 &
  24.50 &
  18.11 &
  52.49 \\
\begin{tabular}[c]{@{}l@{}}Mann-Whitney, U stat\end{tabular} &
  25336 &
  50339 &
  9208 &
  90515 &
  3578 \\
p-value &
  $5.04 \times 10^{-9}$ &
  $3.84 \times 10^{-9}$ &
  $9.36 \times 10^{-5}$ &
  $1.14 \times 10^{-10}$ &
  $1.14 \times 10^{-6}$
\end{tabular}%
}
\caption{Aggregate sCRPS statistics for the distribution of forecasts classified as reasonable/unreasonable by each model for the M5 product dataset.}
\label{tab:m5-agg}
\end{table*}

\begin{tcolorbox}[simplesummary]
    \textbf{Summary: } LLMs can identify real-world forecasts which are likely to be inaccurate. While the number of unreasonable forecasts varies across LLMs (from 2\% up to 40\%), in all cases the sCRPS of unreasonable forecasts exceeds the sCRPS of reasonable forecasts by a significant margin.
\end{tcolorbox}

\section{Conclusion} 
\label{sec:conclusion}

In this work, we have introduced \texttt{The Forecast Critic}, a multi-modal system to assess whether a newly generated forecast is reasonable or unreasonable, plausible or implausible, based on a visualization plus plain-language instructions. 
We have also evaluated how well several modern LLMs can identify good versus obviously-inaccurate forecasts---an essential capability for building scalable and reliable forecasting systems. 
In our experiments using synthetic time series data, we find that LLMs can reliably detect forecast errors (F1 score $> 0.8$), albeit slightly below the performance of expert human evaluators. 
We also demonstrate that our system can effectively use additional text information about promotional covariates, highlighting the promise of LLMs customizing their critiques to incorporate additional context when exogenous shocks arise. 
These results also highlight shortcomings of the LLMs.
For example, LLMs excel at detecting misaligned trends and vertical translations, but they struggle to identify stretched or compressed periodicity.
We also conduct experiments using real-world time series from the M5 dataset with probabilistic forecasts from the Chronos model. We show that unreasonable forecasts have higher sCRPS values than reasonable ones, indicating that the system is detecting unusually inaccurate forecasts. 

Overall, our findings underscore the potential and shortcomings of LLMs as automated forecast auditors. While their current performance remains slightly below that of expert human evaluators, \texttt{The Forecast Critic} framework provides a scalable and reasonable approach to forecast auditing. In future work, we plan to extend this work to include a wider variety of modern LLMs, expand our treatment of additional covariate information, and evaluate our performance on additional real-world time series forecast datasets.

\newpage 

\bibliography{references}

\newpage 
\appendix
\onecolumn 

\section{Definition of statistical metrics}
\label{appendix:stat-met}

\subsection{Weighted F1-Score}
\label{sec:subsec:appendix:f1}

The weighted F1 score is given by
\begin{equation}
\text{Weighted F1 Score} = \sum_{i=1}^{N} \left( \frac{n_i}{\sum_{j=1}^{N} n_j} \cdot F1^{(i)} \right) \nonumber \,, 
\label{eq:weighted-f1}
\end{equation}
where $F1^{(i)}$ is the $F1$ score for the $i$th class, $n_i$ is the number of elements in each class, and $N$ is the total number of classes.

\subsection{Scaled Continuous Ranked Probability Score (sCRPS)} \label{sec:subsec:appendix:scrps}

sCRPS is a normalized version of the standard CRPS, where the total CRPS is divided by the sum of the absolute values of the true time series. Formally, for a given true time series $y$ and a set of forecast quantile predictions $\hat{Y}$, it is defined as 

\begin{equation}
\text{sCRPS}(y, \hat{Y}) = \frac{\sum_{t \in \mathcal{T}_{\text{forecast}}} \text{CRPS}(y_{t}, \hat{Y}_{t})}{\sum_{t \in \mathcal{T}_{\text{forecast}}} |y_{t}|}
\label{eq:scrps} \nonumber \,,
\end{equation}
where $t \in \mathcal{T}_\text{forecast}$ is the number of forecasted points.

To compute the CRPS, we use a Riemann integral approximation technique that uniformly averages the quantile loss over a discrete set of quantiles. The CRPS for a single observation is given by:

\begin{equation}
\text{CRPS}(y, \hat{Y}) = 2 \int_0^1 \text{QL}_q \left(y, F^{-1}_{\hat{Y}}(q) \right) \, dq \nonumber \,, 
\label{eq:crps}
\end{equation}
where we use $q \in  \{0.1, 0.2, ..., 0.9\}$ as the set for the discrete quantiles, $F^{-1}_{\hat{Y}}(q)$ to indicate the predicted quatile at $q$,  and the quantile loss function $\text{QL}_q(\cdot, \cdot)$ is defined as:

\begin{equation}
\text{QL}_q(y, \hat{y}) = q \cdot (y - \hat{y})^+ + (1 - q) \cdot (\hat{y} - y)^+ \nonumber \,,
\label{eq:quantile-loss}
\end{equation}
where \( (x)^+ = \max(x, 0) \).

\section{Additional details on the construction of the synthetic dataset 
}
\label{appendix:sec:dataset}

For our synthetic dataset, we construct unique time series via
\begin{align}
    y(t) = \sum_{i=1}^n w_i \cdot b_i(s_i(t + \delta_i))\,,  \label{eq:ts-main} \quad t \in \mathcal{T}\,,  \nonumber 
\end{align}
where $n \sim \mathcal{U}\{1, 2, 3, 4\}$ is the number of base functions to include, and for all $i \in [1, n]$, we have $w_i \sim \mathcal{U}(0.5, 2.0)$ as an output scaling weight, $s_i \sim \mathcal{U}(0.5, 2.0)$ as an input scaling weight, $\delta_i \sim \mathcal{U}(0, 4)$ as an input shift, and $b_i \sim \mathcal{U}\{1, 2, ..., 14\}$ as the base functions defining the time series, given in Table \ref{tab:basis-fcns} below. 
\begin{table*}[h]
\centering
\begin{tabular}{l|l|l}
ID & base function name & Expression $b_i(t)$                                                                                                                                               \\ \hline
1  & Gaussian wave       & $5 e^{-0.00005 (t-6)^2} \sin(0.5 t)$                                                                                                                              \\
2  & Linear cos          & $0.3 + 0.5t + 0.2 \cos(10t)$                                                                                                                                      \\
3  & Linear              & $0.3 + 0.5 t$                                                                                                                                                     \\
4  & Sin                 & $\sin(4t)$                                                                                                                                                        \\
5  & Sinc                & $10 (t+\epsilon_\text{threshold})^{-1} \sin(5t) $                                                                                                                                \\
6  & Beat                & $\sin(t) \sin(5t)$                                                                                                                                                \\
7  & Sigmoid             & $(1+e^{-4t})^{-1}$                                                                                                                                                \\
8  & Log                 & $\log(1+t)$                                                                                                                                                       \\
9  & Sin scaled       & $4(t+1)\sin(5(t+1)+4)$                                                                                                                                            \\
10 & Square              & $3t^2$                                                                                                                                                            \\
11 & Step                & $\mathscr{H}(t-3, 1)$                                                                                                                                                       \\
12 & Multistep           & \begin{tabular}[c]{@{}l@{}}$0.2 \mathscr{H}(t-1, 1) + 0.3 \mathscr{H}(t-2.5, 1)$ \\ $- 0.1 \mathscr{H}(t-4, 1) + 0.4\mathscr{H}(t-5.5, 1)$ \\ $-0.3\mathscr{H}(t-7, 1) + 0.2\mathscr{H}(t-8.5, 1) + 0.1 \mathscr{H}(t-9.5, 1)$\end{tabular} \\
13 & Chirp               & $\sin(10t^2)$                                                                                                                                                     \\
14 & Sawtooth            & $2\cdot (t \cdot \pi^{-1} - \lceil 0.5 + t\cdot \pi^{-1} \rceil)$                                                                                                
\end{tabular}%
\caption{Various base functions used to construct the synthetic time series dataset. $\mathscr{H}$ corresponds to the Heaviside function and $\epsilon_{\text{threshold}}$ is a numerical stability threshold set to $10^{-10}$. }
\label{tab:basis-fcns}
\end{table*}


\section{Synthetic experiments under controlled perturbations
: Additional results and perturbation construction}
\label{appendix:perturbations}

\subsection{Perturbation types and construction} \label{appendix:subsec:perturbation-constrcution}

\begin{figure*}
    \centering
    \includegraphics[width=0.8\textwidth]{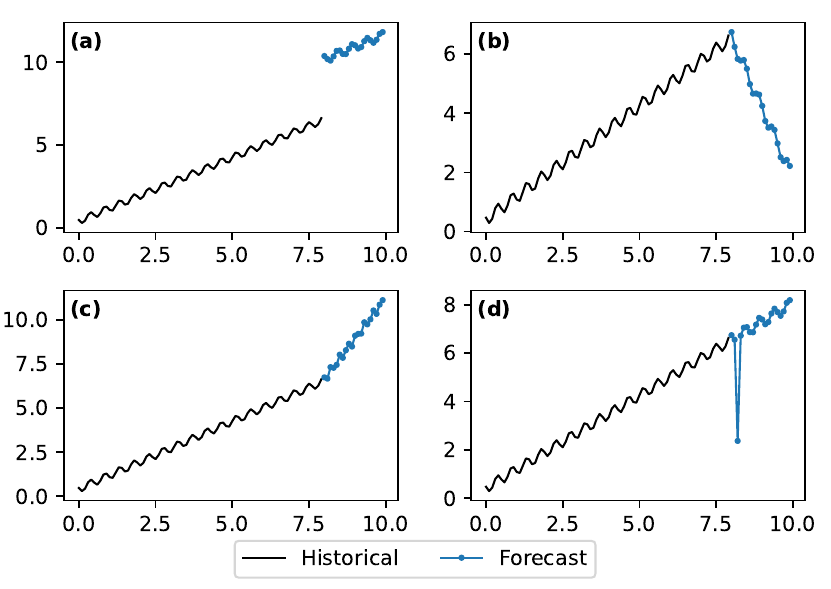}
    \caption{Examples of various perturbations on a synthetic time series dataset. (a) shows the translate up perturbation, (b) shows the trend modification, (c) shows the stretch modification, and (d) has the forecast containing an additional spike. In this case, $(\mathcal{T}_{\text{historical}} = [0, 8]$ and \(\mathcal{T}_{\text{forecast}} = (8, 10]\) where \(\Delta t = 0.1\). Additionally, the hyperparameters for each perturbation are set as \(\omega=0.5\), \(\beta=-3\), \(\alpha=3\), \(\gamma=0.5\), and \(N_{\text{spikes max}} = 3\).}
    \label{fig:perturbations}
\end{figure*}

In this study, we consider four types of point-wise perturbations: translations, trend modifications, additional spiking, and stretching of the signal. Each perturbation simulates a different kind of temporal or structural anomaly. 
We provide visual examples of each type of modification in Figure \ref{fig:perturbations}. Here, we define a general framework, and then we describe each perturbation mathematically:

Consider the discrete signal representation as in  \eqref{eq:ts-main},
\begin{equation}
y(t), \quad t \in \mathcal{T} = \{t_0 + k \cdot \Delta t \mid k = 0, 1, \dots, N\}, \nonumber 
\end{equation}
where \(\Delta t\) is the sampling interval, and the series is divided into historical data \(y(t)\) for \(t \in \mathcal{T}_{\text{historical}}\) and forecast data \(y(t)\) for \(t \in \mathcal{T}_{\text{forecast}}\).

Following this formulation, each perturbation is given as follows.

\begin{itemize}
\item
\textbf{Vertical Shift (Shift scale \(\omega\)).}  
Given the original time series \( y(t) \), the forecast data is perturbed by adding a vertical shift \(\omega\) multiplied by the forecast mean:
\begin{equation}
y_{\text{vertical shift}}(t) := y(t) + \omega \times \frac{1}{|\mathcal{T}_{\text{forecast}}|}\sum_{\theta \in \mathcal{T}_{\text{forecast}}} y(\theta), \quad t \in \mathcal{T}_{\text{forecast}}. \nonumber
\end{equation}

\item
\textbf{Linear Trend Modification (Slope Scaling \(\beta\)).} 
Assume the forecast data can be decomposed into a linear trend plus residuals:
\begin{equation}
y(t) = m t + b + r(t), \quad t \in \mathcal{T}_{\text{forecast}}, \nonumber
\end{equation}
where \(m\) and \(b\) denote the slope and intercept estimated from the forecast data, and \(r(t)\) represents residuals.

The perturbation rescales the slope by \(\beta\):
\begin{equation}
m' = \beta m. \nonumber
\end{equation}

To maintain continuity at \(t_c = t^{(0)}_{\text{forecast}}\), the intercept is adjusted to satisfy:
\begin{equation}
y(t_c) = m' t_c + b' + r(t_c). \nonumber
\end{equation}

Solving for the new intercept gives:
\begin{equation}
b' = y(t_c) - m' t_c - r(t_c). \nonumber
\end{equation}

The resulting perturbed series is:
\begin{equation}
y_{\text{trend modification}}(t) = m' t + b' + r(t). \nonumber
\end{equation}

\item
\textbf{Time Stretching (Multiplicative Factor \(\alpha\)).} 
The time axis is resampled with a new step size \(\alpha \cdot \Delta t\), stretching (\(\alpha > 1\)) or compressing (\(\alpha < 1\)) the signal:
\begin{equation}
\mathcal{T}' = \{t_0 + k \cdot (\alpha \cdot \Delta t)\}, \quad y'(t') = f(t'), \quad t' \in \mathcal{T}', \nonumber
\end{equation}
where \(f(t)\) is the generating signal for the time series (i.e., the right-hand side of \eqref{eq:ts-main}).

Further, to ensure continuity at the forecast boundary:
\begin{align}
y_{\text{time stretch}}(t') &= y'(t') + \delta, \nonumber \\
\delta &:= y(t_{\text{forecast}}^{(0)}) - y'(t_{\text{forecast}}^{(0)}). \nonumber
\end{align}

\item
\textbf{Random Spikes (Modifier \(\gamma\), Max Count \(N_{\text{spikes max}}\)).} 
Randomly inject \( n \sim \text{Uniform}\{1, \dots, N_{\text{max}}\} \) spikes into the forecast data at positions \(\mathcal{T}_{\text{spike}} \subset \mathcal{T}_{\text{forecast}}\), where \(|\mathcal{T}_{\text{spike}}| = n\):
\begin{equation}
y_{\text{spikes}}(t) =
\begin{cases}
y(t) + \epsilon, & t \in \mathcal{T}_{\text{spike}}, \\
y(t), & \text{otherwise},
\end{cases} \nonumber
\end{equation}
where the spike magnitude \(\epsilon\) is defined as
\begin{equation}
\epsilon = \pm \gamma \cdot \max_{t \in \mathcal{T}_{\text{forecast}}} y(t), \nonumber
\end{equation}
with the sign chosen independently and uniformly at random for each spike.
\end{itemize}

\subsection{Example of LLM responses for various synthetic perturbations} 
\label{sec:appendix:llm_perturb_examples}

\begin{tcolorbox}[imgllmbox]
    \begin{minipage}{0.48\textwidth}
        \centering
        \includegraphics[width=\linewidth]{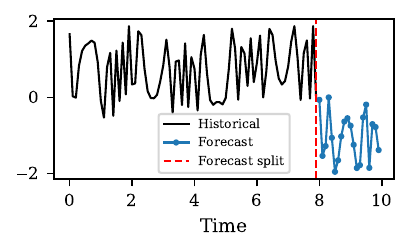}  
    \end{minipage}
    \hspace{5pt} 
    \begin{minipage}{0.48\textwidth}
        \begin{tcolorbox}[quotebubble]
            Looking at the image, I can see that the historical data (black line) shows a highly oscillatory pattern with values typically ranging between -1 and 2, with most values staying in the positive range. However, the forecast (blue line) shows oscillations primarily in the negative range (between approximately -2 and 0), representing a significant downward level shift from the historical pattern. While the forecast does capture the oscillatory nature of the data, the substantial shift downward makes it inconsistent with the historical behavior.

            \par \vspace{0.5em}
            
            The forecast is unreasonable. The forecast has captured the oscillating pattern but incorrectly shifted the entire prediction downward into a predominantly negative range when the historical data mostly fluctuated in the positive range.
        \end{tcolorbox}
    \end{minipage}
\end{tcolorbox}
\noindent \begin{minipage}{\textwidth}
\captionof{figure}{
Example of an LLM response corresponding to a shift misalignment, as shown in Figure \ref{fig:perturbations}(a). The LLM correctly recognizes that the transition point is poor, and hence it flags the forecast as unreasonable.}
\label{fig:perturb-example-1}
\end{minipage}

\begin{tcolorbox}[imgllmbox]
    \begin{minipage}{0.48\textwidth}
        \centering
        \includegraphics[width=\linewidth]{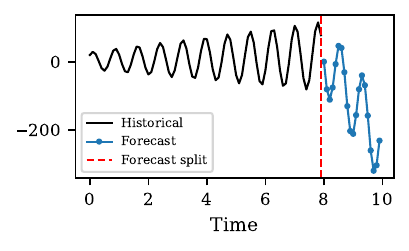}  
    \end{minipage}
    \hspace{5pt} 
    \begin{minipage}{0.48\textwidth}
        \begin{tcolorbox}[quotebubble]
 Looking at this image, I can see that the historical data (black) shows a clear oscillating pattern with gradually increasing amplitude over time. The forecast (blue) continues the oscillation pattern but with a dramatically larger amplitude than what would be expected from the historical trend.
\par \vspace{0.5em}
 The forecast is unreasonable.
\par \vspace{0.5em}
The forecast is unreasonable because while it maintains the cyclical nature of the data, it shows a sudden and extreme jump in amplitude that doesn't align with the more gradual amplitude increase shown in the historical pattern.
        \end{tcolorbox}
    \end{minipage}
\end{tcolorbox}
\noindent \begin{minipage}{\textwidth}
\captionof{figure}{
Example of an LLM response corresponding to a trend misalignment, as shown in Figure \ref{fig:perturbations}(b). The LLM correctly recognizes that the amplitude differs from the historical trend, and hence it flags the forecast as unreasonable.}
\label{fig:perturb-example-2}
\end{minipage}

\begin{tcolorbox}[imgllmbox]
    \begin{minipage}{0.48\textwidth}
        \centering
        \includegraphics[width=\linewidth]{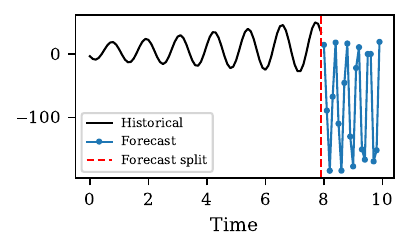}  
    \end{minipage}
    \hspace{5pt} 
    \begin{minipage}{0.48\textwidth}
        \begin{tcolorbox}[quotebubble]
Looking at the image, I can see that the historical data (black line) shows a consistent sinusoidal pattern with regular oscillations between approximately $+25$ and $-25$. However, the forecast (blue line) dramatically changes in amplitude, showing extreme negative values down to around $-150$ while maintaining similar frequency. This represents a significant and unexplained departure from the established pattern.
\par \vspace{0.5em}
 The forecast is unreasonable.
\par \vspace{0.5em}
The forecast is unreasonable because it dramatically increases the amplitude of oscillations and introduces extreme negative values that aren't consistent with the historical pattern, while the frequency remains similar.
        \end{tcolorbox}
    \end{minipage}
\end{tcolorbox}
\noindent \begin{minipage}{\textwidth}
\captionof{figure}{
Example of an LLM response corresponding to a compression perturbation, as shown in Figure \ref{fig:perturbations}(c). The LLM correctly recognizes that the forecast oscillation frequency increase beyond reason, and hence it flags the forecast as unreasonable.}
\label{fig:perturb-example-3}
\end{minipage}

\begin{tcolorbox}[imgllmbox]
    \begin{minipage}{0.48\textwidth}
        \centering
        \includegraphics[width=\linewidth]{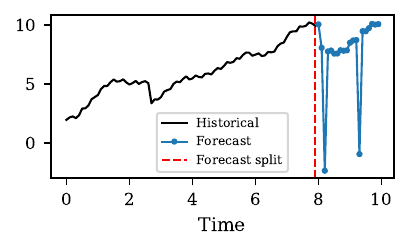}  
    \end{minipage}
    \hspace{5pt} 
    \begin{minipage}{0.48\textwidth}
        \begin{tcolorbox}[quotebubble]
Looking at the image, I can see that while the forecast (blue line) continues the overall upward trend from the historical data (black line), it introduces extreme volatility that isn't present in the historical pattern. The historical data shows a relatively smooth upward progression, but the forecast contains dramatic drops (particularly one that goes negative around time 9.5) and sharp fluctuations that aren't justified by any previous patterns in the data.
\par \vspace{0.5em}
 The forecast is unreasonable. The forecast introduces extreme volatility and dramatic drops (including one below zero) that aren't consistent with the relatively smooth upward pattern shown in the historical data.
        \end{tcolorbox}
    \end{minipage}
\end{tcolorbox}
\noindent \begin{minipage}{\textwidth}
\captionof{figure}{
Example of an LLM response corresponding to the presence of additional spikes, as shown in Figure \ref{fig:perturbations}(d). The LLM correctly recognizes that there are unnecessary spikes in the forecast, and hence it flags the forecast as unreasonable.}
\label{fig:perturb-example-4}
\end{minipage}

 
\section{Exogenous Feature Injection
: Additional results and experimental details} 
\label{sec:appendix:holidays}

\subsection{Example of the input images for the promotional input scenarios, considered an example of the prompt with Exogenous Feature Injection as plain text.} 
\label{sec:appendix:holiday_prompts}

\begin{figure}[h]
    \centering
    \includegraphics[width=0.9\textwidth]{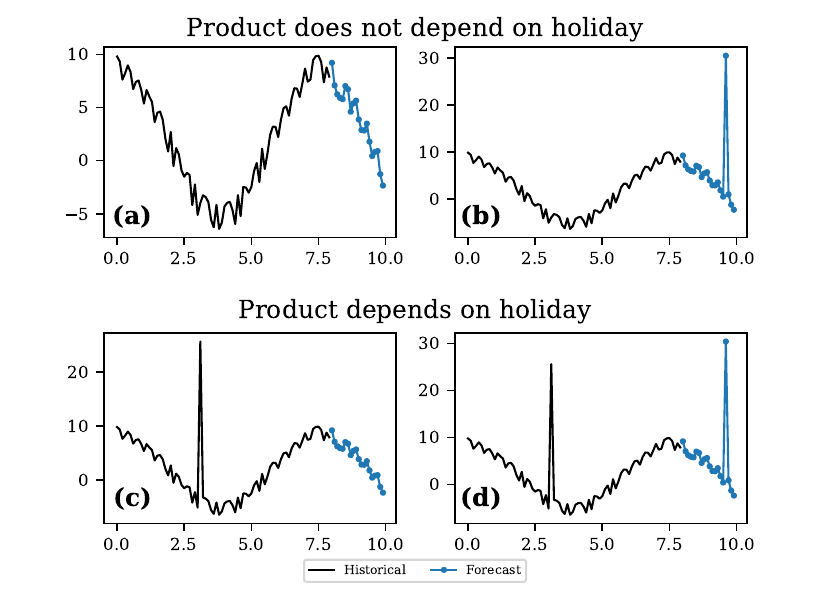}
    \caption{Examples of the four different promotional scenarios considered 
    in the Exogenous Feature Injection section of the main text. 
    The top row considers the case where the product is not dependent on the holiday, but the user provides holiday information. Hence, the true label for (a) is reasonable and for (b) is unreasonable. The bottom row considers the case where the product depends on the holiday. Hence, the true label for (c) is unreasonable, since the forecast is missing the demand spike, and for (d) is reasonable.}
    \label{fig:holiday-examples}
\end{figure}

\begin{tcolorbox}[promptbox]
\begin{minipage}{0.48\textwidth}
    You are shown an image of historical data (in black) and a forecast (in blue). 
    Based on the historical trend, assess whether the forecast is reasonable.
    
    A reasonable forecast should generally follow the same direction and capture any seasonal 
    trends if there are any. Note, there is a holiday at $t=0.321$ in the historical
    and a second holiday at time $t=8.47$ in the forecast that may affect the demand. 
    
    Please provide a brief explanation (1–2 sentences) justifying your decision. 
    Then present your final answer using one of the following options, wrapped in <answer> tags:
    
    <answer> 1 </answer> — The forecast is reasonable.  
    <answer> 2 </answer> — The forecast is unreasonable.   
\end{minipage}%
\hfill
\begin{minipage}{0.48\textwidth}
\centering
\includegraphics[width=\linewidth]{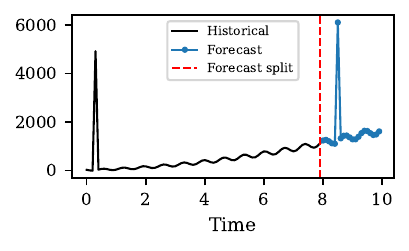}
\end{minipage}
\end{tcolorbox}
\begin{minipage}{\textwidth}
\captionof{figure}{An example prompt for the exogenous signal scenarios 
in the Exogenous Feature Injection section of the main text. 
In this scenario, the forecast correct accounts for holiday spikes. }
\label{fig:holiday_example_prompt}
\end{minipage}

\subsection{Examples of LLM responses for all Exogenous Feature Injection experiment scenarios} \label{sec:appendix:holiday:feature_examples}

\begin{tcolorbox}[imgllmbox]
    \begin{minipage}{0.48\textwidth}
        \centering
        \includegraphics[width=\linewidth]{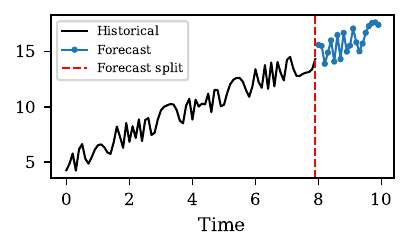}  
    \end{minipage}
    \hspace{5pt} 
    \begin{minipage}{0.48\textwidth}
        \begin{tcolorbox}[quotebubble]
   Looking at the image, I can see that the forecast (blue line) continues the same upward trend established by the historical data (black line), and importantly, it maintains the regular oscillation patterns present in the historical data. The forecast also appears to accommodate the expected holiday impact at $t=8.24$ similarly to how the historical data showed variations around the holiday at $t=4.29$.
        \end{tcolorbox}
    \end{minipage}
\end{tcolorbox}
\noindent \begin{minipage}{\textwidth}
\captionof{figure}{
Example of an LLM holiday response for scenario (a) 
in the Exogenous Feature Injection section of the main text. 
The LLM correctly recognizes that since the the product does not have a significant spike during the holiday in the forecast history, there should not be a spike in the forecast.}
\label{fig:exogenous-example-1}
\end{minipage}

\begin{tcolorbox}[imgllmbox]
    \begin{minipage}{0.48\textwidth}
        \centering
        \includegraphics[width=\linewidth]{paper_figures/holiday_scenario_2_example-1.pdf}  
    \end{minipage}
    \hspace{5pt} 
    \begin{minipage}{0.48\textwidth}
        \begin{tcolorbox}[quotebubble]
  Looking at the image, I can see that the historical data (black line) shows a clear cyclical pattern with decreasing amplitude over time. The forecast (blue line) maintains the cyclical pattern but doesn't continue the dampening trend, instead showing relatively stable oscillations. Additionally, the holiday spike at $t=9.82$ in the forecast appears significantly larger than the comparable holiday spike in the historical data at $t=1.88$, which doesn't align with the established pattern. These inconsistencies suggest the forecast doesn't properly capture the historical trend characteristics.
        \end{tcolorbox}
    \end{minipage}
\end{tcolorbox}
\noindent \begin{minipage}{\textwidth}
\captionof{figure}{
Example of an LLM holiday response for scenario (b) 
in the Exogenous Feature Injection section of the main text. 
The LLM correctly recognizes that the spikes at $t=9.82$ is much larger then the comparable spike in the history, and hence it flags the example.}
\label{fig:exogenous-example-2}
\end{minipage}

\begin{tcolorbox}[imgllmbox]
    \begin{minipage}{0.48\textwidth}
        \centering
        \includegraphics[width=\linewidth]{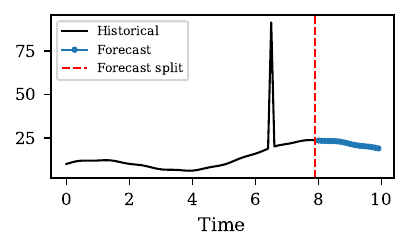}  
    \end{minipage}
    \hspace{5pt} 
    \begin{minipage}{0.48\textwidth}
        \begin{tcolorbox}[quotebubble]
    Looking at the image, I can see the historical data shows a relatively flat pattern with a massive spike at $t=6.53$ (the first holiday), reaching a value around $80$. However, the forecast doesn't show any comparable spike at $t=8.87$ where the second holiday occurs, instead maintaining a smooth, gradually declining trend around the $20-25$ value range. Given that holidays typically create similar patterns in time series data, the forecast fails to capture this critical seasonal effect from the first holiday.
        \end{tcolorbox}
    \end{minipage}
\end{tcolorbox}
\noindent \begin{minipage}{\textwidth}
\captionof{figure}{
Example of an LLM holiday response for scenario (c) 
in the Exogenous Feature Injection section of the main text. 
The LLM correctly recognizes that the spike is not comparable at $t=8.87$, and hence it correctly flags the forecast.}
\label{fig:exogenous-example-3}
\end{minipage}

\begin{tcolorbox}[imgllmbox]
    \begin{minipage}{0.48\textwidth}
        \centering
        \includegraphics[width=\linewidth]{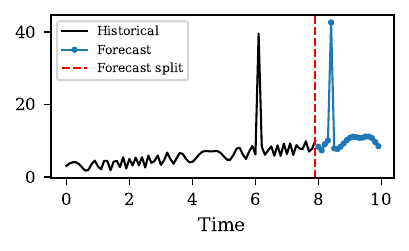}  
    \end{minipage}
    \hspace{5pt} 
    \begin{minipage}{0.48\textwidth}
        \begin{tcolorbox}[quotebubble]
  Looking at the image, I can see that the historical data shows a relatively stable baseline with a significant spike at $t=6.09$ (the first holiday). The forecast appropriately captures a similar spike pattern at the second holiday ($t=8.43$) and maintains a reasonable baseline level consistent with the recent historical trend. The magnitude of the holiday spike in the forecast is proportionate to the historical holiday spike, and the forecast continues the general pattern observed in the historical data.
        \end{tcolorbox}
    \end{minipage}
\end{tcolorbox}
\noindent \begin{minipage}{\textwidth}
\captionof{figure}{
Example of an LLM holiday response for scenario (d) 
in the Exogenous Feature Injection section of the main text. 
The LLM correctly recognizes that the forecast maintains the trend of the historical data and accuractely captures the spikes given at the holiday promotional dates.}
\label{fig:exogenous-example-4}
\end{minipage}

\section{Real-world forecasts from foundation models: 
Additional experimental details and results}

\subsection{Example of probabilistic forecasting prompt input}

\begin{tcolorbox}[promptbox]
\begin{minipage}{0.48\textwidth}
You are shown an image of historical data (in black) and a forecast (in blue). 
Your task is to assess whether the forecast appears visually reasonable.

A reasonable forecast should generally follow the same direction as the historical trend 
and reflect any clear seasonal patterns, if present.

IMPORTANT: Only label a forecast as unreasonable if there is an obvious and significant mismatch — 
for example, if the forecast goes in the opposite direction of the trend, ignores strong seasonal 
patterns, or shows extreme jumps that are not supported by the historical data. 

Minor deviations or slight over/underestimates are acceptable and should still be considered reasonable.

Please provide a brief explanation (1–2 sentences) justifying your decision. 
Then present your final answer using one of the following options, wrapped in <answer> tags:

<answer> 1 </answer> — The forecast is reasonable.  
<answer> 2 </answer> — The forecast is unreasonable.  

If you find the forecast unreasonable, clearly explain what makes it obviously inconsistent.
\end{minipage}%
\hfill
\begin{minipage}{0.48\textwidth}
\centering
\includegraphics[width=\linewidth]{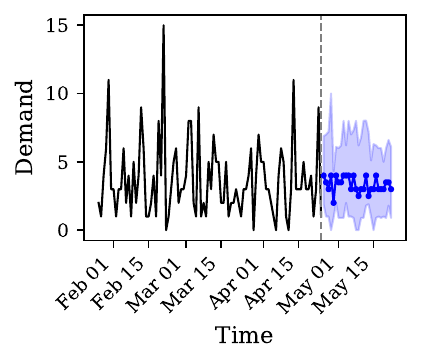}
\includegraphics[width=0.9\linewidth]{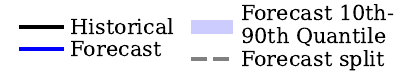}
\end{minipage}
\end{tcolorbox}
\begin{minipage}{\textwidth}
\captionof{figure}{An example prompt for the M5 dataset problems. 
}
\label{fig:m5_example_prompt}
\end{minipage}

\subsection{Additional anecdotes of flagged forecasts on the M5 dataset with Chronos}
\begin{tcolorbox}[imgllmbox]
    \begin{minipage}{0.48\textwidth}
        \centering
        \includegraphics[width=\linewidth]{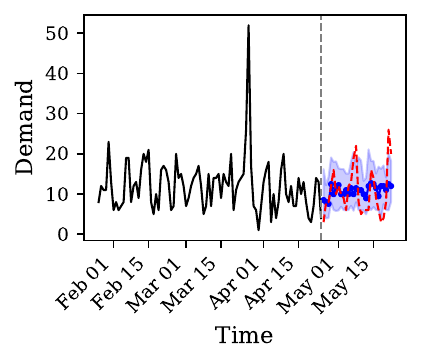}  
    \end{minipage}
    \hspace{5pt} 
    \begin{minipage}{0.48\textwidth}
        \begin{tcolorbox}[quotebubble]
        Looking at this image, I can see historical data (black line) that shows considerable volatility with frequent oscillations between approximately 5 and 20, plus one major spike reaching around 50. The blue forecast line appears much smoother and predicts values around 10-15 with minimal fluctuation.
        
        \par \vspace{0.5em}
        
        The forecast is unreasonable because it fails to capture the significant volatility pattern that's clearly present in the historical data. The historical series shows regular ups and downs with occasional large spikes, but the forecast appears much smoother without accounting for this established pattern of fluctuation.
        \end{tcolorbox}
    \end{minipage}
    \centering 
    \includegraphics[width=0.8\linewidth]{paper_figures/legend-1.pdf}
\end{tcolorbox}
\noindent \begin{minipage}{\textwidth}
\captionof{figure}{
Example of a \emph{incorrectly} flagged forecast by the Claude 3.7 Sonnet. This error is due to the fact that the single spike in the historical data caused the LLM to overbias the expected volatility, despite the majority of the data following the forecasted~pattern.}
\label{fig:m5-ad-2}
\end{minipage}

\begin{tcolorbox}[imgllmbox]
    \begin{minipage}{0.48\textwidth}
        \centering
        \includegraphics[width=\linewidth]{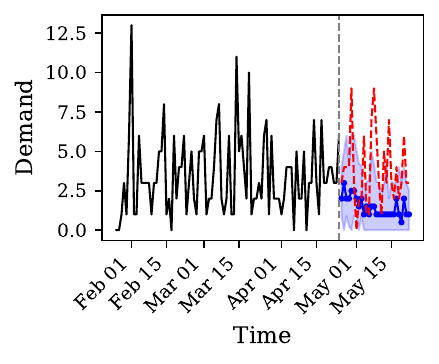}  
    \end{minipage}
    \hspace{5pt} 
    \begin{minipage}{0.48\textwidth}
        \begin{tcolorbox}[quotebubble]
         Looking at the image, I can see historical data (black line) that shows significant volatility with frequent spikes between 0 and approximately $12.5$ units. The blue forecast line that follows, however, appears much flatter and lacks the characteristic volatility pattern shown in the historical data. 
         
         \par 
         \vspace{0.5em}
         While the general level of the forecast might be within a reasonable range, it fails to reflect the significant up and down movements that are consistently present throughout the historical period. The forecast is unreasonably smooth compared to the highly volatile historical data. The historical series shows frequent spikes and significant amplitude in its movements, but the forecast shows minimal volatility, suggesting a fundamental mismatch in capturing the time series behavior.
        \end{tcolorbox}
    \end{minipage}
    \centering 
    \includegraphics[width=0.8\linewidth]{paper_figures/legend-1.pdf}
\end{tcolorbox}
\noindent \begin{minipage}{\textwidth}
\captionof{figure}{
Example of another correctly flagged forecast by LLM forecast critic.}
\label{fig:m5-ad-3}
\end{minipage}

\end{document}